%% file: main.tex
\documentclass{article}

\usepackage[preprint]{neurips_2026}
\usepackage[utf8]{inputenc}
\usepackage[T1]{fontenc}
\usepackage{graphicx}
\usepackage{subcaption}
\usepackage{booktabs}
\usepackage{amsmath}
\usepackage{amssymb}
\usepackage{mathtools}
\usepackage{amsthm}
\usepackage{amsfonts}
\usepackage[dvipsnames]{xcolor}
\usepackage{array}
\usepackage{colortbl}
\usepackage{url}
\usepackage{tikz}
\usetikzlibrary{arrows.meta, positioning, calc}

\usepackage{placeins}
\usepackage{color}
\usepackage{soul}
\usepackage{xspace}
\usepackage{hyperref}
\hypersetup{
    colorlinks=true,
    citecolor=TolMutedBlue,%
    linkcolor=TolMutedGreen,%
    urlcolor=TolMutedPurple,%
}
\usepackage[capitalize,noabbrev]{cleveref}
\crefname{section}{Sec.}{Sec.}
\crefname{thm}{Thm.}{Theorem}
\crefname{appendix}{App.}{Appendices}
\crefname{algorithm}{Alg.}{Algorithms}
\crefname{equation}{Eq.}{Eqs.}
\crefname{figure}{Fig.}{Figs.}
\crefname{prop}{Prop.}{Props.}
\creflabelformat{equation}{#2\textup{#1}#3} %

\theoremstyle{plain}

\theoremstyle{definition}

\theoremstyle{remark}

\newcounter{promptnum}

\usepackage[disable,textsize=tiny]{todonotes}

\input{defns.tex}

\newif\ifprint

\ifprint

\else

\fi

\newcommand{\cja}[1]{\todo[color=purple!40]{CJA: #1}}

\definecolor{mplblue}{HTML}{0015bc}
\definecolor{mplorange}{HTML}{8c000f}

\definecolor{TolMutedBlue}{HTML}{332288}
\definecolor{TolMutedGreen}{HTML}{117733}
\definecolor{TolMutedPurple}{HTML}{AA4499}

\title{
  Quantifying the Agreement Between Data-Influence\\
  and Data-Similarity to Understand LLM Behavior
}

\author{
  \textbf{Christopher J. Anders}$^{1}$ \quad
  \textbf{Henrique Da Silva Gameiro}$^{2*}$ \quad
  \textbf{Nico Daheim}$^3$ \\[4pt]
  \textbf{Mohammad Emtiyaz Khan}$^{1,4}$ \\[1em]
  $^1$RIKEN Center for Advanced Intelligence Project, Tokyo, Japan \\
  $^2$Section of Computer Science, EPFL Lausanne, Switzerland \\
  $^3$Ubiquitous Knowledge Processing Lab (UKP Lab), \\
  Department of Computer Science, Technical University of Darmstadt \\
  National Research Center for Applied Cybersecurity ATHENE, Germany \\
  $^4$TU Darmstadt \& Hessian Center for AI (hessian.AI), Darmstadt, Germany \\[1em]
  $^1$\texttt{\{christopher.anders,emtiyaz.khan\}@riken.jp} \\
  $^2$\texttt{henrique.dasilvagameiro@gmail.com} \quad
  $^4$\texttt{nico.daheim@tu-darmstadt.de} \\
}

\begin{document}

\maketitle

\let\thefootnote\relax\footnotetext{* This work is based in part on a Master's thesis completed during an
internship at RIKEN AIP.}

\begin{abstract}
  One way to understand LLM behavior is to trace its output back to the training data. Two types of measures are commonly used for output tracing: data-similarity and data-influence. The former is cheaper while the latter is believed to be more accurate. Even though many works have compared them for ground-truth tasks, no such comparisons exist for output tracing.
  Here, we fill this gap and precisely quantify the commonalities and differences between the two measures.~We do this by first ranking the training documents according to each measure and then computing the overlap between the two rankings.
  Our main finding is that the two rankings agree significantly, but there is an \emph{asymmetry}
  between them: The top documents of data-similarity are assigned more consistent ranks by data-influence than the other way around.~This result is valid across a range of experiments involving OLMo2-1B, Qwen3-1.7B, LlaMa3.2-1B,
  Gemma3-1B, and GPT2.~We exploit the asymmetry to obtain a favorable cost-accuracy trade-off by using the costly data-influence to refine the results of data-similarity.
\end{abstract}

\section{Introduction}

Understanding working-mechanisms of large language models (LLMs) is important to address serious concerns regarding their behaviour, for example, those regarding hallucinations and biases \citep{xiao2021hallucination,zheng2023judging,wu2025style}, breach of privacy \citep{barbera2025ai}, and copyright infringement \citep{karamolegkou2023copyright,chang2023speak}.
Yet, discovering the causes of such issues is challenging due to the size of the model and
complexity of its training process. The answer is hidden somewhere among the large training corpus, buried in the millions of parameters, and arises due to various architectural and algorithmic choices made during training.

\begin{figure}[t!]
  \begin{center}
      \includegraphics[width=0.9\linewidth]{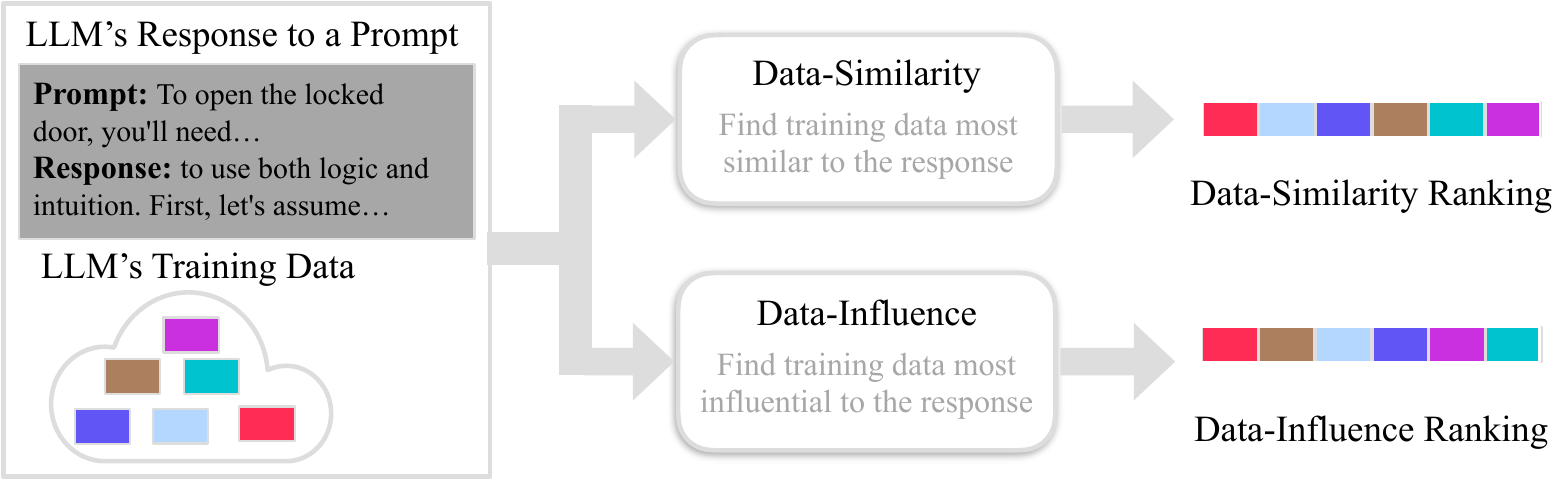}
    \caption{One way to understand an LLM's behavior is to probe it with a prompt and trace its
    response back to its training data. Two types of tracing methods have been used in the
    literature: data-similarity and data-influence. Both return a ranking over the training data.
    We compare these rankings to understand where they agree and disagree, and which should be preferred over the other.}
    \label{fig:fig1}
  \end{center}
\end{figure}

Recent efforts bypass these difficulties by treating LLMs as information-retrieval systems, where
we first probe them with prompts and then trace their responses back to training data
(\cref{fig:fig1}).
For instance, \citet{liu2025olmotrace} propose `OLMo-Trace' which uses \emph{data-similarity} measures based on string matching to trace the outputs of OLMo-2 (32B parameters) to its multi-trillion-token training data. Several other proposals also reported similar success in retrieving relevant documents, for example, by using BM25~\citep{robertson2009probabilistic,kamphuis2020which} and
InfiniGram~\citep{liu2024infinigram}.
Such applications clearly demonstrate the effectiveness and scalability of data-similarity measures to better understand LLMs trained on massive data sets.

An alternate approach is to use \emph{data-influence}, which measures the influence of training data over a response.
This can be estimated, for example,
by using the model's predictions \citep{paul2021deep}, gradients \citep{pruthi2020estimating}, and/or Hessians \citep{koh2017understanding}.
Several works have used this approach for a variety of purposes, for example, to understand LLM generalization \citep{grosse2023studying},
training-data attribution \citep{barshan2020relatif} and data-valuation~\citep{choe2024what}.
It has also been used to improve LLM training and fine-tuning procedures
\citep{thakkar2023self,xia2024less,joaquin2024in2core}.
In general, data-influence approaches are believed to be more accurate and, but data-similarity measures are often \emph{much} cheaper.

Even though many works have compared the two measure by their performance on ground-truth
tasks~\citep{akyurek2022towards,chang2024scalable}, no such comparisons exist for output tracing.
Output-tracing is conceptually closer to search engines,
where an \emph{absolute} ground-truth for rankings rarely exist.
Therefore, past comparisons based on ground-truth are not directly useful to compare the two measures for output tracing.
It thus remains unclear how similar or different
these measures are for output-tracing, or whether they can be combined in some way to get the best of both worlds.
For example, this can be useful for OLMo-Trace, which currently only uses data-similarity
measures. It is not known whether data-influence measures can improve such systems and whether it
is worth to put in that extra cost.

\begin{figure*}[t!]
  \begin{center}
    \centerline{
      \includegraphics{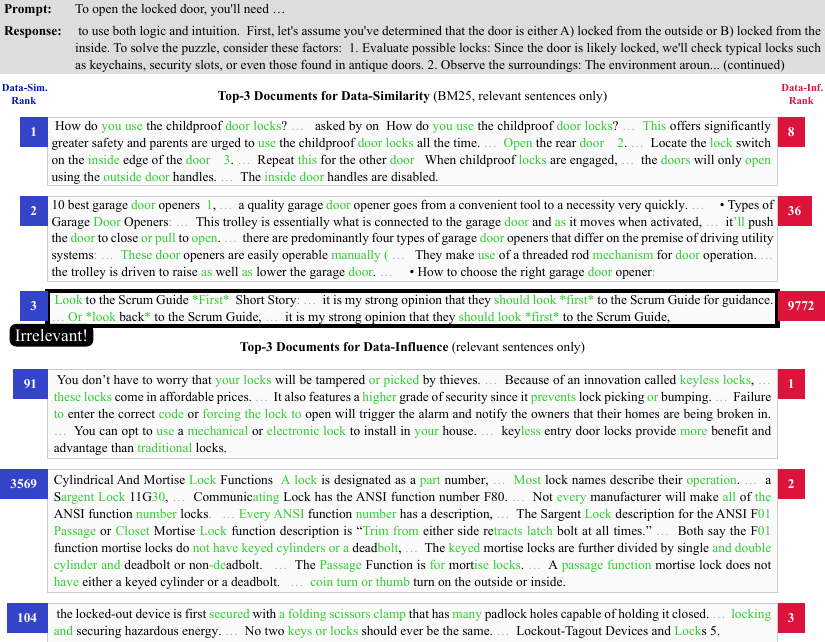}
    }
    \caption{Similarities and differences in the rankings assigned by data-similarity and
    data-influence. Top: The prompt-response query for OLMo2-1B on `opening locked-doors'.
    Below: The top-3 documents traced by each approach, with rankings (left/right) and
    relevant words/phrases highlighted in green.
    \textbf{All documents are highly relevant except for the third-ranked
    document by data-similarity, which data-influence ranks 9,772nd}.
    On average, data-influence assigns better rankings, reflecting that its rankings are more
    predictive of the rankings of data-similarity than the other way around.
    More queries can be found in~\cref{app-qualitative}.
    }
    \label{fig:fig2}
  \end{center}
\end{figure*}

In this paper, we fill this gap and precisely quantify the commonalities and differences
between data-similarity and data-influence for output tracing. We do this by comparing their
rankings through ranking metrics that compute overlap at a certain \emph{depth}, for example, considering a depth of top-100 rankings \citep{webber2010similarity}. 
Such overlaps reveals more nuanced details about the relationship between the two measures than the approaches that rely on ground-truth tasks.
Our main finding is that the two rankings agree significantly, but there is an \emph{asymmetry}
between them. For instance, the top-20 documents found by
data-similarity are assigned more consistent ranks by data-similarity than the other way around (see \cref{fig:fig2,fig-stem}). This shows that data-influence can better predict the ranking of data-similarity than the other way around.
As an example, below is a summary of our results for OLMo2-1B:
\vspace{-6pt}
\begin{enumerate}
  \setlength{\itemsep}{-0.8pt}
  \item Agreement: 11\% of the top-100 documents of the two measures are the same.
  \item Asymmetry: 34\% of the top-100 documents of data-similarity fall within
     the top-1000 documents of data-influence, while this falls down to 28\% when considering the opposite.
  \item Consistency: better influence measures agree more with data-similarity, for example, a 1\% agreement with gradient-based influence goes up to $11\%$ when Hessians are also used.
\end{enumerate}
\vspace{-4pt}
Similar results are obtained on six different LLMs (see \cref{fig-symm}), including
Llama-3.2-1B and Qwen3-1.7B.
Overall, these results highlight the complementary strengths of the two measures and suggest
that a good cost-accuracy trade-off can be achieved by combining them.
Specifically, we can first retrieve a candidate set using the cheaper data-similarity measure and
then refine them using the more expensive data-influence measure; see \cref{fig-auc-cross}.

\section{Understanding LLMs by Output Tracing}

LLM understanding is challenging due to their scale but recent work has demonstrated the practicality and effectiveness of output-tracing methods for extremely large models. OLMoTrace builds upon the data-similarity measure, such as infini-gram and BM25, to trace outputs of the OLMo-2-32B-Instruct model back to its dataset consisting of multi-trillion tokens. Data-Influence approaches have also been applied to large problems, for example, to Llama3-8B-Instruct
over the 1B-token dataset in \citet{choe2024what} and LLMs with 52B parameter in~\citet{grosse2023studying}. We will now describe some of these approaches and justify the choice of methods we compare in this paper.

\subsection{Data Similarity}
  \label{sec-representation-similarity}

  Data-similarity measures for text documents essentially measure the similarity in terms of
  frequency of co-occurrence of words and, more generally, `terms'. For example, given a vocabulary
  with $M$ terms, we can simply count how many times a term occurred in a document and use it to represent the document. More precisely, for the $i$-th document, we can construct a feature
  vector $\vphi_i$ of length $M$ whose $t$'th entry contains the count of the $t$'th term. This is the classical `bag of words' model. More sophisticated models use bi-gram or $n$-grams to represent documents, where we count pairs of two or $n$ terms.
  To avoid the counts being dominated by commonly-used words, it is common to divide the term-frequency (TF) by the inverse of document frequency (IDF), giving rise to the popular TF-IDF features.
  Given the feature vectors, the data-similarity measure is straightforwardly defined by using an inner product, for instance, $\mathcal{S}(i,j) = \vphi_i^\top \vphi_j$.

  A popular alternative is the Best-Matching (BM) algorithm~\citep{robertson2009probabilistic}, whose BM25 version has been used in several works on LLMs, most recently in OLMoTrace \citep{liu2025olmotrace}. The BM25 algorithm employs an `asymmetric' inner product \citep{wu2011learning} where the usual bag-of-words vector is used for $\vphi_i$ but the vector $\vphi_j$ is normalized by the IDFs in a specific way;
  a detailed expression is given in \cref{eq:bm25} for the Lucene version of BM25
  \citep{kamphuis2020which}. In many works, BM25 is used as a baseline and often shows good
  performance, for instance, fact-tracing \citep{akyurek2022towards,chang2024scalable}.
  \citet{liu2025olmotrace} used it to create the final ranking of the document and found to have,
  as the authors quote, `fairly high agreement with human judgement on topical relevance'. We will
  therefore use BM25 for data-similarity.

  We note that OLMoTrace also uses a more recent faster alternative called Infini-gram
  \citep{liu2024infinigram} to efficiently count queries and retrieve matching documents in massive
  text corpora, which for OLMo-2 contains trillions of tokens. This, along with BM25, enables to trace outputs verbatim back to the full training data and enables real-time traces. OLMoTrace clearly demonstrates that data-similarity methods are practical and useful tools to understand LLMs
  via output tracing.

\subsection{Data Influence}
Data-influence aims to estimate the influence of training data over the model parameters and its
predictions. It was original proposed for supervised problems \citep{cook1979influential,
koh2017understanding}, but has been repurposed for LLMs by considering a set of input tokens or a
whole document as training data examples \citep{grosse2023studying,choe2024what}. At test time, we want to quantify the influence of such documents on a query, say, a prompt-response pair. 
The influence of a text-document on a query can be obtained by comparing the changes in the response when we replace the original LLM by another LLM trained \emph{without} that document (or a set of tokens).

We can write this more precisely by denoting the response to the $i$'th query of an LLM with parameter $\vparam$ by $\vf_i(\vparam)$.
Then, to define the influence of the $j$'th document, we consider another LLM with parameter
$\vparam_{\minus j}$ obtained by removing the $j$'th document from the training data. The influence
of $j$'th document on the $i$'th query can be defined as
\begin{equation}
   \influence(i,j) = \| \vf_i(\vparam) -\vf_i(\vparam_{\minus j}) \|,
   \label{eq:influence}
\end{equation}
where $\|\cdot\|$ is a norm. Instead of removal, we can also measure the influence under other
perturbation, for example, by reweighting the document differently. The type and size of
perturbation can also vary according to the problem.

The above influence-measure provides a definitive answer to important \emph{what-if} scenarios, but
expensive model retraining is infeasible. Fortunately, cheap
approximations can be used instead. For example, instead of full retraining, we can simply take one
gradient or Newton step to estimate the influence. This can be easily implemented with back-prop and has been extensively used for LLMs.
Below, we show two popular strategies using gradient vectors $\vg_i$ and Hessian $\vH$,
\begin{equation}
   \influence_g(i,j) = \vg_i^\top \vg_j
   \qquad
   \influence_h(i,j) = \vg_i^\top \vH^{-1} \vg_j.
   \label{eq:two_types_of_influence}
\end{equation}
A popular example of gradient-based estimators is TracIn \citet{pruthi2020estimating} where $\vg_i$
are gradients of the loss function and $\influence_g(i,j)$ are averaged over intermediate
training checkpoints. For LLMs, these may also be averaged over tokens.
A popular example of Newton-style estimator is the classical Influence-Function
\citep{koh2017understanding, grosse2023studying}, which is so commonly used that often the word
`influence' is confused with 'influence functions'.
Most works on LLMs do in fact use the Newton-style estimator \citep{grosse2023studying, choe2024what}.

The accuracy and cost of data-influence techniques directly depend on the type of approximations
used \citep{nickl2024memory, hong2025better}. For instance,
Newton-style estimators are expected to yield better estimates, but they require an expensive estimation of the
Hessian.
Gradient-based estimators are cheaper as they skip the expensive Hessian, but are expected to yield worse estimates.
Similarly, better estimators can be obtained by using fine-tuning, which is a common technique for unlearning too \citep{liu2025rethinking}.
In general, the accuracy of influence estimators should increase with more compute, which is an attractive property of data-influence measures.

In this paper, we will consider using both $\influence_g$ and $\influence_h$.
Specific choices of these estimators are discussed in \cref{sec-methods}.

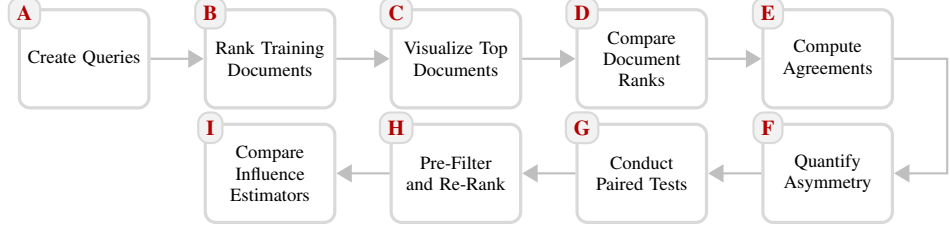
\begin{figure*}[t]
    \centering
    \begin{tikzpicture}[
            node distance=5pt and 20pt,
            block/.style={
                rectangle,
                rounded corners=4pt,
                draw=black!15,
                very thick,
                text width=1.5cm,
                minimum height=1.3cm,
                align=center,
                inner sep=3pt,
                font=\scriptsize,
            },
            header/.style={
                rectangle,
                rounded corners=4pt,
                draw=black!15,
                fill=black!5,
                thick,
                inner sep=3pt,
                font=\footnotesize\bfseries,
                text=red!70!black,
                anchor=center
            },
            arrow/.style={
                -{Triangle[scale=1.2]},
                thick,
                black!25
            }
        ]

        \node [block] (a)             {Create Queries};
        \node [block, right=of a] (b) {Rank Training Documents};
        \node [block, right=of b] (c) {Visualize Top Documents};
        \node [block, right=of c] (d) {Compare Document Ranks};
        \node [block, right=of d] (e) {Compute Agreements};
        \node [block, below=of e] (f) {Quantify Asymmetry};
        \node [block, left=of f] (g) {Conduct Paired Tests};
        \node [block, left=of g] (h) {Pre-Filter and Re-Rank};
        \node [block, left=of h] (i) {Compare Influence Estimators};

        \foreach \n/\L in {a/A, b/B, c/C, d/D, e/E, f/F, g/G, h/H, i/I} {
            \node [header] at ($(\n.center)!0.9!(\n.north west)$) {\L};
        }

        \draw [arrow] (a) -- (b);
        \draw [arrow] (b) -- (c);
        \draw [arrow] (c) -- (d);
        \draw [arrow] (d) -- (e);

        \draw [arrow] (e.east) -- ++(20pt, 0pt) |- (f.east);

        \draw [arrow] (f) -- (g);
        \draw [arrow] (g) -- (h);
        \draw [arrow] (h) -- (i);

        \end{tikzpicture}
    \caption{Experimental workflow to compare the rankings of data-influence and data-similarity.}
    \label{fig:workflow}
\end{figure*}

\newcommand\makestemtableA{
  \begin{minipage}[c]{0.75in}
    \scriptsize
    \setlength{\arraycolsep}{2pt}
    \setlength{\tabcolsep}{2pt}
    \begin{tabular}{|>{\raggedleft\arraybackslash}m{0.30in}|>{\raggedleft\arraybackslash}m{0.35in}|}
      \specialrule{\heavyrulewidth}{0pt}{0pt}
      \rowcolor{mplblue!20}
      \textbf{Top-20 of Sim} & \textbf{Rank assigned by Inf} \\
        \specialrule{\lightrulewidth}{0pt}{0pt}
        1 & 1 \\
        2 & 6 \\
        3 & 2 \\
        4 & 25 \\
        5 & 57 \\
        6 & 3 \\
        7 & 12 \\
        8 & 5 \\
        9 & 13 \\
        10 & 14 \\
        11 & \hl{160} \\
        12 & 51 \\
        13 & 8 \\
        14 & 46 \\
        15 & 40 \\
        16 & 74 \\
        17 & 17 \\
        18 & \hl{155} \\
        19 & \hl{201} \\
        20 & 33 \\
        \specialrule{\heavyrulewidth}{0pt}{0pt}
    \end{tabular}
  \end{minipage}
}
\newcommand\makestemtableB{
  \begin{minipage}[c]{0.75in}
    \scriptsize
    \setlength{\arraycolsep}{2pt}
    \setlength{\tabcolsep}{2pt}
    \begin{tabular}{|>{\raggedleft\arraybackslash}m{0.30in}|>{\raggedleft\arraybackslash}m{0.35in}|}
      \specialrule{\heavyrulewidth}{0pt}{0pt}
      \rowcolor{mplorange!20}
      \textbf{Top-20 of Inf} & \textbf{Rank assigned by~Sim} \\
        \specialrule{\lightrulewidth}{0pt}{0pt}
        1 & 1 \\
        2 & 3 \\
        3 & 6 \\
        4 & \hl{107} \\
        5 & 8 \\
        6 & 2 \\
        7 & 86 \\
        8 & 13 \\
        9 & \hl{6291} \\
        10 & 25 \\
        11 & 77 \\
        12 & 7 \\
        13 & 9 \\
        14 & 10 \\
        15 & \hl{253} \\
        16 & \hl{243} \\
        17 & 17 \\
        18 & \hl{1090} \\
        19 & \hl{181} \\
        20 & \hl{176} \\
        \specialrule{\heavyrulewidth}{0pt}{0pt}
    \end{tabular}
  \end{minipage}
}

\subsection{Data-Similarity vs Data-Influence}

Having described both approaches, we make a final point regarding an important distinction between
the two measures, because it is easy to confuse them. To some, the influence measures $\influence_g(i,j)$
can also be seen as a data-similarity measures where the feature vector is replaced by the
gradient. In fact, it is common in the literature to refer to influence measures as similarity
metric and vice-versa~\citep{akyurek2022towards,sun2025enhancing,guo2021fastif}. The line is further blurred
when model-embeddings are used as features for measures based on cosine similarity \citep{singla2023simple}.

To avoid such confusion, we will use the following definitions throughout this paper:
\vspace{-6pt}
\begin{itemize}
  \setlength{\itemsep}{-0.4pt}
  \item `Data-Similarity' is reserved for methods that build the feature purely based on the data
    and \emph{never} use the LLM outputs. Specifically, we will use BM25.~We do not consider embedding-based
    methods, which are expected to be better. These are not used in systems such as OLMo-Trace
    probably because they are computationally expensive compared to BM25.
   \item `Data-Influence' is reserved for the gold-standard counterfactual shown in \cref{eq:influence}. Influence measures are referred to those that aim to estimate \cref{eq:influence}. Influence-Function (IF) refers to a specific influence-measures $\influence_h$ that uses the Newton step as shown in \cref{eq:two_types_of_influence}.
\end{itemize}
\vspace{-6pt}

As discussed earlier, little work has been done to precisely quantify the complementary strengths of the two approaches. Previous studies, such as those on fact
tracing~\citep{akyurek2022towards,chang2024scalable}, do not provide a definite conclusion for
output-tracing where the goal is to find training documents related to the outputs. We will
precisely quantify the similarities and differences between the two approaches for tracing LLM outputs.

\section{Quantifying Agreement Between Data-Similarity \& Data-Influence}\label{sec-methods}

  \begin{figure*}[t!]
    \begin{center}
      \begin{subfigure}[b]{1.45in}
        \resizebox{1.45in}{!}{
        \makestemtableA
        \makestemtableB
        }
        \caption{}
        \label{fig:stem-table}
      \end{subfigure}
      \hfill
      \begin{subfigure}[b]{4in}
        \includegraphics{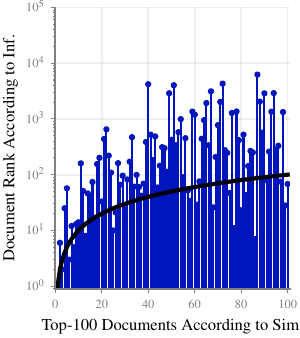}
        \includegraphics{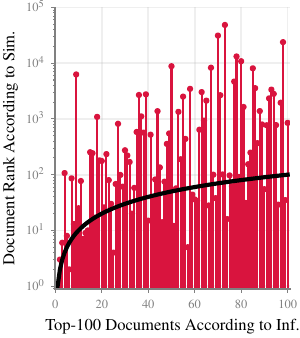}
        \caption{}
        \label{fig:stem-plot}
      \end{subfigure}
      \caption{
        This figure shows both agreement between Inf and Sim and that Inf ranks the top documents according to Sim superior compared to
        how Inf ranks the top documents according to Inf.
        The two tables on the left (a) show the descrepency between the setups in the top-20,
        where ranks beyond the top-100 are highlighted in yellow.
        The two plots on the right (b) visualize the top-100 on the horizontal axis,
        and the ranks assigned by the other measure on the vertical axis in logarithmic scale.
        The solid black line indicates the best possible match between the measures.
        The results are shown for the query:
        \emph{``A cat's favorite activity is usually...''} (see App.~\cref{tbl-prompts}).
      }
      \label{fig-stem}
    \end{center}
  \end{figure*}

In this section, we describe our method to quantify commonalities and differences between data-similarity (Sim) and data-influence (Inf).
For Sim, we use the Lucene~\citep{kamphuis2020which} version of BM25.
For Inf, we primarily use the Hessian-based EK-FAC estimator.
This influence estimator is quite expensive, which puts a limit on the number of documents and
prompts we can use for influence computation. Using the whole training data is infeasible due to
the matrix inverse.
Therefore, we are forced to choose a subset of 100,000 documents and use only 100 prompts to
estimate influence. Computing EK-FAC for one prompt on 100,000 training documents on OLMo2-1B
required about one hour on 16 NVIDIA V100 GPUs. We therefore limited the number of prompts to 100,
which is the highest our computational budget allowed (about 18 days for all LLMs);
see \cref{sec:limitations} for further explanations and \cref{app:details} for more details.
We show comparisons on the following LLMs: \emph{OLMo2-1B}, \emph{Qwen3-1.7B}, \emph{LlaMa3.2-1B},
\emph{Gemma3-1B}, \emph{GPT2-medium}, and \emph{GPT2-small}.

We now give a step-by-step description of our workflow given in \cref{fig:workflow}.

\textbf{A. Create Queries:}
We generate 100 prompts of diverse topics and varying ambiguity using
Gemini~\citep{gemini2025gemini} and manually verify them; see \cref{fig:fig2} for an example and
\cref{tbl-prompts} in the appendix for a full list of all queries.
Each prompt is completed by each of the pre-trained LLMs.
The LLM's response is concatenated to the prompt, forming a \emph{query}.
The queries represent trials in our empirical evaluations.

\textbf{B. Rank Training Documents:}
We compute Sim scores between each training document and each query.
Then, we compute absolute Inf scores for each training document to the LLM's
autoregressive loss on each query.
Finally, we sort the training documents according to the scores of each of the measures.
For each query, this yields one ranking according to Sim, and one ranking according to Inf.
See \cref{fig:stem-table} for an example of the two rankings.

\textbf{C. Visualize Top Documents:}
For the top-3 documents of each measure,
we compute the per-token contribution to each score as
described in~\cref{eq:two_types_of_influence_llm,eq:bm25} in the appendix.
We then show only sentences with an average per-token score of at least 10\% of the highest
per-token score in the document,
and further highlight all tokens whose \emph{individual} score is at least this value.
See \cref{fig:fig2} for an example.

\emph{Results:} In \cref{fig:fig2}, we see some commonalities between the measures, where
both measures mostly retrieve documents related to the query.
But there are also differences.
For instance, Sim retrieves an irrelevant document about SCRUM.
For another query in the appendix (\cref{fig-sentencetoken-47-bm25,fig-sentencetoken-47-ekfac}),
we see that the top-3 match very closely,
but for yet another query in the appendix (\cref{fig-sentencetoken-79-bm25,fig-sentencetoken-79-ekfac}),
Sim and Inf do not resemble each other at all.
In the following, we will precisely quantify the agreement between the two measures through
their rankings.

\textbf{D. Compare Document Ranks:}
We define the top-100 as the relevant documents (this number can be chosen
differently based on the setup and use-case).
For each measure (data-similarity and data-influence), we
(1) pick the relevant documents,
(2) then sort them by their rank,
and (3) finally annotate each document with its rank assigned by the other measure.
We show the top-20 in two tables in \cref{fig:stem-table} and highlight ranks assigned by the other
measure that exceed 100.
We then show the top-100 in two stem plots in \cref{fig:stem-plot}.

\emph{Results:}
In \cref{fig:stem-table}, we see agreement between Sim and Inf:
Firstly, Sim and Inf assign the same document to rank 1.
Secondly, the majority of documents in the top-20 of Sim are also in the top-100 of Inf.
This likewise holds the other way around.
But we also observe asymmetry: Seven documents in the top-20 of Inf are outside the
top-100 of Sim (highlighted), but only three documents in the top-20 of Sim are outside the
top-100 of Inf.
The stem plots in \cref{fig-stem}
show a similar trend:
The stems grow with increasing ranks,
indicating that the top-100 of both measures roughly agree compared to the rank assigned by the other measure.
But for the top-100 according to Sim, the stems are overall lower than for Inf, revealing a
potential asymmetry in their agreement.

\textbf{E. Compute Agreements:}
We precisely quantify the agreement between the top-$d$ according to each measures by counting the number of
documents they have in common.
Dividing this number by $d$ yields a score between 0 and 1, appropriately known as the
\emph{agreement} at depth $d$~\citep{webber2010similarity}; see \cref{fig-agreement} for an example and \cref{eq-agreement} in the appendix.
By visualizing the agreement at all rank depths $d\in\{1,...,100000\}$, we can measure at which ranks they agree most.
Here, high agreement in the top-ranks indicates that the measures retrieve similar examples.
We show the mean and variance of the agreement over all queries for each of the six LLMs in \cref{fig-symm}.
Further, we can represent high agreement in the top ranks with a single score using rank-biased
overlap~\citep{webber2010similarity},
which we use to compare the agreement between queries in \cref{fig-query-rbo} in the appendix.

\begin{figure}[t]
  \begin{subfigure}[t]{0.48\textwidth}
    \vspace{0pt}
    \begin{center}
      \small
      \setlength{\arraycolsep}{2pt}
      \setlength{\tabcolsep}{4pt}
      \resizebox{\textwidth}{!}{
      \begin{tabular}{|l|l|l|c|}
        \specialrule{\heavyrulewidth}{0pt}{0pt}
        \rowcolor{gray!20}
        $d$ & Ranking 1 & Ranking 2 & Agreement \\
          \specialrule{\lightrulewidth}{0pt}{0pt}
          1 &
          $\begin{pmatrix} 1 \end{pmatrix}$ &
          $\begin{pmatrix} 7 \end{pmatrix}$ &
          0.000 \\

          2 &
          $\begin{pmatrix} 1 & 2 \end{pmatrix}$ &
          $\begin{pmatrix} 7 & 3 \end{pmatrix}$ &
          0.000 \\

          3 &
          $\begin{pmatrix}
              1 & 2 & 3
          \end{pmatrix}$ &
          $\begin{pmatrix}
              7 & 3 & 1
          \end{pmatrix}$ &
          0.667 \\

          4 &
          $\begin{pmatrix}
              1 & 2 & 3 & 4
          \end{pmatrix}$ &
          $\begin{pmatrix}
              7 & 3 & 1 & 4
          \end{pmatrix}$ &
          0.500 \\

          5 &
          $\begin{pmatrix}
              1 & 2 & 3 & 4 & 5
          \end{pmatrix}$ &
          $\begin{pmatrix}
              7 & 3 & 1 & 4 & 2
          \end{pmatrix}$ &
          0.800 \\

          6 &
          $\begin{pmatrix}
              1 & 2 & 3 & 4 & 5 & 6
          \end{pmatrix}$ &
          $\begin{pmatrix}
              7 & 3 & 1 & 4 & 2 & 5
          \end{pmatrix}$ &
          0.833 \\

          7 &
          $\begin{pmatrix}
              1 & 2 & 3 & 4 & 5 & 6 & 7
          \end{pmatrix}$ &
          $\begin{pmatrix}
              7 & 3 & 1 & 4 & 2 & 5 & 6
          \end{pmatrix}$ &
          1.000\\

        \bottomrule
      \end{tabular}
      }
      \vspace{2em}
      \caption{}
      \label{fig-agreement}
    \end{center}
  \end{subfigure}
  \hfill
  \begin{subfigure}[t]{0.48\textwidth}
    \vspace{0pt}
    \begin{center}
      \centerline{
        \includegraphics{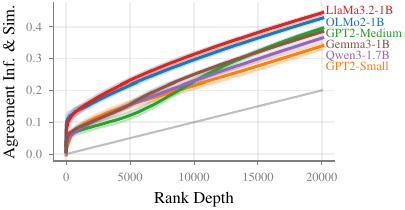}
      }
      \caption{}
      \label{fig-symm}
    \end{center}
  \end{subfigure}
  \vskip -1em
  \caption{
    (a) The \emph{agreement} of two example rankings
    on the same set of size $7$
    with increasing depth $d$ (see App.~\cref{eq-agreement}).
    Agreement does not increase monotonically with depth,
    but will always reach 1.0 when the rankings are based on the same set.
    (b) Agreement between Inf and Sim for six LLMs.
    The steep climb of the curves close to rank 1
    indicates agreement in the top ranks.
    The variance is shown over 100 queries.
    The bottom gray line is the expected agreement for unrelated rankings.
    For all 100,000 ranks, see App.~\cref{fig-symm-full}.
  }
  \end{figure}
\emph{Results:}
\Cref{fig-symm} shows the agreement up to rank 20,000.
We observe a sharp increase in the top-tier documents for all LLMs, which indicates some agreement.
Especially in the first 100 examples, all lines rise sharply.
Nevertheless, there is also significant disagreement.
The rank-biased overlap values shown in \cref{fig-query-rbo} in the appendix suggest that the agreement depends on the query:
for some queries there is high, and almost complete agreement.~For other queries, there is no agreement at all.

\newcommand\infsim{Inf$\to$Sim}
\newcommand\siminf{Sim$\to$Inf}

\textbf{F. Quantify Asymmetry:}
To quantify the asymmetry between Sim and Inf,
we measure how predictive their rankings are of each other.
For this, take the top-100 according to each measure and compare how well the other measure ranks
them.
If the ranking assigned by the other measure significantly exceeds 100, then clearly it is not
predictive of those ranks.
If one measure is consistently doing better than the other in such predictions, then we can conclude that
one is more informative of the other, indicating an asymmetric relationship.
To this end, we do the following prediction tasks:
\vspace{-6pt}
\begin{enumerate}
  \setlength{\itemsep}{-0.4pt}
  \item \textbf{\infsim:} We predict the ranking of Sim using Inf, that is, we take the top-100 documents of Sim and obtain their respective rankings according to Inf
  \item \textbf{\siminf:} We do the opposite: take the top-100 documents of Inf and obtain their respective rankings according to Sim
\end{enumerate}%
\vspace{-6pt}
As before, we show the mean and variance over our 100 queries for each LLM.

  \begin{figure*}[t]
    \begin{center}
    \begin{subfigure}[b]{4.05in}
      \centerline{
        \includegraphics{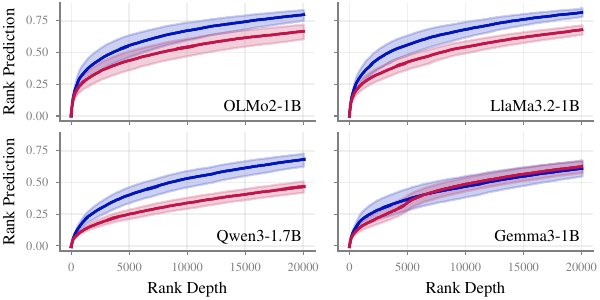}
        \llap{
          \includegraphics{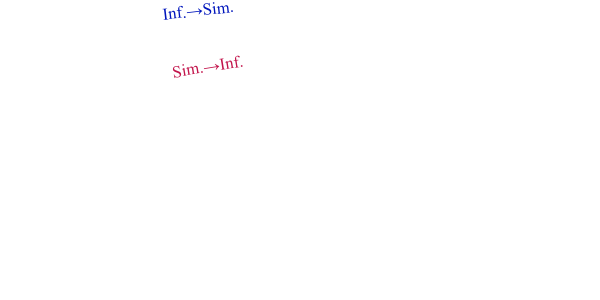}
        }
      }
      \caption{}
      \label{fig-asymm}
    \end{subfigure}
    \hfill
    \begin{subfigure}[b]{1.41in}
      \centerline{
        \includegraphics{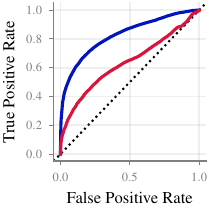}
        \llap{
          \includegraphics{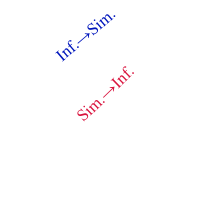}
        }
      }
      \vspace{2em}
      \caption{}
      \label{fig-auc-cross}
    \end{subfigure}
      \caption{
        Results suggesting that Inf is better at predicting Sim than the other way around.
        (a) Rank prediction of the top-100 of Sim (BM25) using Inf (KFAC),
        and vice-versa for four LLMs.
        For all 100,000 ranks and six LLMs, see \cref{fig-asymm-full}.
        (b) Pre-filtering using Sim and refining using Inf and vice versa, evaluated using the
        receiver-operator characteristic on OLMo2-1B.
        See step \textbf{H} for details.
      }
    \end{center}
  \end{figure*}
  \emph{Results:}
  In \cref{fig-asymm}, we see that
  across all rank depths $d$ beyond the symmetric case $d=100$, \infsim~(upper blue curve) appears
  consistently higher than \siminf~(lower red curve).
  For instance, for OLMo2-1B, about 80\% of the top-$100$ examples according to Sim are included in the
  top-$20000$ examples according to Inf.
  But only about 65\% of the top-$100$ according to Inf are included in the
  top-$20000$ according to Sim.
  Gemma3-1B is an exception above rank 5000, where \siminf~catches up, but the top-ranks are
  of primary interest.
  These results suggests Inf is more predictive of Sim than the other way around, demonstrating their asymmetric relationship.

\textbf{G. Conduct Paired Tests:}
We test for statistically significance between the prediction tasks in step \textbf{F} above
using the Wilcoxon signed rank test.
Based on the measurements of each of the two tasks, \infsim~and \siminf,
we formulate the null- and alternative hypotheses as follows.
Let $D_d$ be the median difference between \infsim~and \siminf~at rank depth $d$.
\begin{align}\label{eq:hypo}
  H_0:&~ \text{The median difference } D_d \text{ is less or equal zero.}\\
  H_a:&~ \text{The median difference } D_d \text{ is larger zero.}
\end{align}
The null hypothesis assumes
that Inf is less or equally predictive of Sim than the other way around.

  \emph{Results:}
  The p-values for the null hypothesis at each rank depth $d$ are visualized in \Cref{fig-p-value}
  in the appendix.
  We find that the p-values calculated under the assumption of the null hypothesis are far below
  0.05 for all rank depths except close to rank depths 100 and 100,000.
  The agreement between the two rankings at and around a depth of 100 is (close-to) symmetric,
  therefore one cannot be more predictive of the other.
  Similarly, at rank depths close to the maximum rank,
  almost all examples are included in both rankings,
  which explains the high agreement in any ranking method.
  In between, we can safely reject the null Hypotheses and therefore conclude convincingly that
  Inf is more predictive of Sim than the other way around.%

  \textbf{H. Pre-Filter and Refine:}
  Exploring all training examples with data-influence is prohibitive in practice.
  With pre-filtering, we can avoid computing influence on the whole dataset.
  However, there are no works that study the quality of such pre-filtering.
  In the following, we provide a proof-of-concept for pre-filtering with Sim and refining with
  Inf, and compare it to the (unrealistic) inverse setup.
  For each measure, we do the following:
  We retrieve the top-100 documents for each query, and label them \emph{positive}.
  We randomly choose 100 documents from the training set for each query, and label them \emph{negative}.
  Then, we use the respective other measure to assign scores to the 100 retrieved and 100 random documents of each query.
  This leads to the same relationship as in step \textbf{F}:%
  \vspace{-6pt}
  \begin{enumerate}
    \setlength{\itemsep}{-0.4pt}
    \item \textbf{\infsim:} Retrieve with Sim and score with Inf
    \item \textbf{\siminf:} Retrieve with Inf and score with Sim
  \end{enumerate}%
  \vspace{-6pt}
  Using the scores and labels for the documents of all queries,
  we plot the receiver-operating characteristic (ROC) curve in~\cref{fig-auc-cross}.

  \emph{Results:} For \siminf, we observe an area under the ROC curve (AUC) of 0.63.
  This means that a considerable portion of the top-100 documents retrieved by Inf may be missed by Sim.
  The opposite direction \infsim~seems to be more viable, with an AUC of 0.83.
  The outcome is thus in agreement with findings in step \textbf{F}.

  \textbf{I. Compare Influence Estimators:}
  The use of cheaper influence estimators may impact our previous observations.
  Therefore, we investigate whether our results hold across different influence estimators.
  We repeat steps \textbf{E} and \textbf{F} with the following influence estimators:
  (1) Hessian-based EK-FAC (KFAC), (2) its diagonal approximation (DIAG)~\citep{george2018fast},
  and (3) the gradient-based estimator in \cref{eq:two_types_of_influence} (GRAD).
  KFAC is a better but more expensive estimator than DIAG, which itself is better but more
  expensive than GRAD.
  We show results for OLMo-2-1B.

  \begin{figure}[t]
    \begin{subfigure}{0.49\textwidth}
      \begin{center}
        \centerline{
          \includegraphics{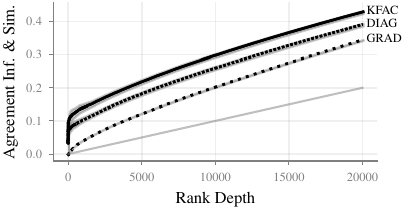}
        }
        \caption{}
        \label{fig-symm-multiple}
      \end{center}
    \end{subfigure}\begin{subfigure}{0.49\textwidth}
      \begin{center}
        \centerline{
          \includegraphics{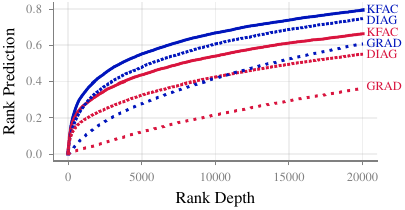}
          \llap{
            \includegraphics{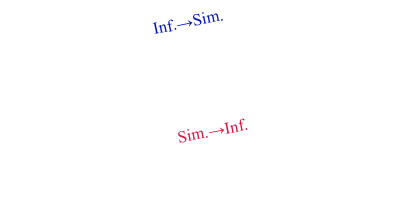}
          }
        }
        \caption{}
        \label{fig-asymm-multiple}
      \end{center}
    \end{subfigure}
    \vskip -1em
    \caption{
      Results showing the consistency of our observations across various influence estimators on
      OLMo2-1B.
      (a) Agreement between Sim (BM25) and Inf (various).
      The variance is shown over 100 queries.
      The bottom gray line is the expected agreement for unrelated rankings.
      (b) Rank prediction of the top-100 of Sim (BM25) using Inf (various) and vice-versa.
      The variances are hidden to improve readability.
      For all 100,000 ranks and variances, see App.~\cref{fig-symm-multiple-full,fig-asymm-multiple-full}.
    }
  \end{figure}

  \emph{Results:}
  In~\cref{fig-symm-multiple} we observe agreement across all estimators.
  However, agreement appears to consistently increase with better influence estimators,
  which is indicated by the steep increase in the top ranks.
  The same applies to the asymmetry between Sim and Inf: In~\cref{fig-asymm-multiple}, \infsim~(upper red
  curves) appears consistently higher than \siminf~(lower blue curves) across all estimators.
  These results show that the observed (asymmetric) agreement is consistent across influence
  estimators and scales with the estimation quality:
  the better the influence estimator, the more Sim and Inf agree, and the better Inf is at predicting Sim.

\section{Conclusion}

In this work,
we show how to quantify the complementary strengths
of data-influence and data-similarity for tracing LLM outputs.
We achieve this by comparing the rankings assigned by the two measures
to the training documents for a range of models
including OLMo2-1B, Qwen3-1.7B, LlaMa3.2-1B, Gemma3-1B, and GPT2.
Our main finding is that the two rankings agree significantly, but there is an \emph{asymmetry}
between them: The top documents of data-similarity are assigned more consistent ranks by
data-influence than the other way around.
These results are consistent across three common data-influence estimators,
were we observe better estimators to yield higher agreement.
This suggests that a good cost-accuracy trade-off for tracing the output of LLMs can be
achieved by combining data-similarity and data-influence.
In a proof-of-concept, we use data-similarity to find a candidate set of
relevant documents, which we refine using data-influence.
Our work opens up the path for hybrid versions of tools such as OLMo-Trace, where the combined
strengths of data-similarity and data-influence can lead to better understanding of LLMs.
In the future, we wish to evaluate how well such hybrid approaches work
compared to existing ones in real-time applications.

\paragraph{Limitations and Future Work}\label{sec:limitations}

Our work has two primary limitations.
First, due to computational constraints, we limited the number of queries and training documents. Each query requires an expensive influence estimation on \emph{all} training
documents, making it infeasible to consider more queries and training documents given our
computational budget (16 NVIDIA V100 GPUs).
Second, while our set of prompts cover a wide variety of topics, they were not explicitly
balanced or categorized by subjects.
As a consequence, it is unclear to what degree the observed agreement between data-similarity and
data-influence depends on the specific prompt.
These limitations present several promising directions for future work.
First, an important next step is to scale our analysis to larger query and
training document sets. For instance, assuming linear runtime scaling and an acceleration
factor of $1 \times \text{H200} \approx 3.27 \times \text{V100}$ (NVIDIA GPUs), an influence estimation for 1000 prompts on
1 billion training documents would require approximately 20 days on a cluster of 1024 NVIDIA
H200 GPUs.
Second, while our results across different model sizes up to 1.7 billion parameters suggest that
data-similarity and data-influence (asymmetrically) agree irrespective of the number of parameters,
verifying this trend on larger, production-grade models (e.g., 30 billion parameters or more) remains
an important open question.
Third, categorizing prompts by topic could provide a deeper understanding of when cheaper data-similarity is
sufficient versus when more accurate data-influence is necessary.
Such insights could be leveraged at test-time to dynamically predict whether a specific query
requires the use of data-influence.
Finally, it would be highly valuable to investigate the extent to which other data-similarity
measures, such as InfiniGram or semantic embeddings, agree with data-influence.

\section*{Acknowledgements}
CJA and MEK were supported by the Bayes duality project, JST CREST Grant Number
JPMJCR2112. ND was supported by the German Federal Ministry of Research, Technology and Space and
the Hessian Ministry of Higher Education, Research, Science and the Arts within their joint support
of the National Research Center for Applied Cybersecurity ATHENE.

\bibliography{main}
\bibliographystyle{icml2026}

\newpage
\appendix
\crefalias{section}{appendix}

\FloatBarrier
\section{Additional Background}
\FloatBarrier

\subsection{Influence on LLMs}

  The influence $\influence(i,j)$ and its estimators $\influence_g(i,j)$ and $\influence_h(i,j)$
  correspond to the single token predictions $i$ under the removal of a single token loss $j$ in LLMs.
  However, in our experiments, we are interested in the change in
  the generated document $\calA$ under the removal of both individual
  tokens $j$, and full training documents $\calD$.
  For this setting,
  the influence estimators simply become the sum
  over the gradients of the tokens contained in the documents $\calA$ and $\calD$
  \citep{koh2019accuracy,grosse2023studying}, with
  \begin{equation}
    \influence_g(\calA,\calD) = \sum_{i\in\calA}\vg_i^\top \sum_{j\in\calD}\vg_j
     \qquad
     \influence_h(\calA,\calD) = \sum_{i\in\calA}\vg_i^\top \vH^{-1} \sum_{j\in\calD}\vg_j.
     \label{eq:two_types_of_influence_llm}
  \end{equation}
  where we can reduce the training document $\calD$ to a single token
  when we are interested in the token-wise influence scores,
  such as shown in~\cref{fig:fig2}

\subsection{Token-wise and Document-wise BM25}

The ``Lucene''~\citep{kamphuis2020which}  version of BM25~\citep{robertson2009probabilistic},
can be defined as an asymmetric similarity \citep{wu2011learning} with
\begin{align}\label{eq:bm25}
  \text{BM25}(\vq, \vd) &= \phi_Q(\vq)^\top \phi_D(\vd) \\
  \phi_Q(\vq) &= \left\{|\vq|_{w_i}~\right\}_{i=1}^{M} \nonumber\\
  \phi_D(\vd) &=
      \Big\{\frac{\text{IDF}_i \cdot |\vd|_{w_i} }{
        k_1 (
          1 - b + b \frac{|\vd|}{\overline{D}}
        ) + |\vd|_{w_i}
      }
    ~\Big\}_{i=1}^{M}\nonumber\\
  \text{IDF}_i &=
    \log\left(
      1 + \frac{N - |w_i| + 0.5}{|w_i| + 0.5}
    \right),\nonumber
\end{align}
given some vocabulary
$\vw = \left( w_1, w_2, \dots, w_M\right)$
with vocabulary size $M$.
Here, $k_1 \geq 0$ is a hyper-parameter,
$|\vd|_{w_i}$ is the number of times token $w_i$ appears in $\vd$,
$|\vd|$ is the document length of document $\vd$,
$\overline{D}$ is the average document length,
and $|w_i|$ is the number of documents in which word $w_i$ appears at least once.
We obtain the scores for the qualitative visualization in \cref{fig:fig2} and others
through the individual components of $\phi_D(\vd)$ for each term in the document.

\subsection{Rank Correlation}

  In this work, we compare rankings of
  training documents $\vX: \real^{N \times d}$
  obtained by measures of similarity
  $k: \real^{d} \times \real^{d} \to \real$
  to some query $\vq$.
  Formally, we define the rank of training document $\vx_i$
  \begin{align}\label{eq-ranking}
    r_{j,i}(k) = 1 + \sum_{{\substack{{m=1}\\m\neq i}}}^N
      \mathbb{I}\left(
        k(\vq,\vx_i) > k(\vq,\vx_m)
      \right),
  \end{align}
  \cja{add tie-breaker? (secondary rank by index $i<j$)}
  for instance, the document $\vx_i$ with the highest similarity $k(\vq, \vx_i)$
  to query $\vq$ will be assigned rank $r_{j,i}(k) = 1$.

  A classical approach to compare these is the
  Spearman rank correlation, which is defined as the Pearson correlation
  of rankings.
  However, it is not suitable for our analysis, as
  (1) it does not allow comparing sets that contain different documents,
  which is required in our rank prediction experiment,
  (2) it does provide a way to focus only on the top-ranks which we are primarily interested in,
  and
  (3) it is sensitive to the magnitude of the difference between the ranks, causing the bottom ranks to dominate.
  For this reason, we follow \citet{webber2010similarity} and use \emph{agreement} in our
  evaluation of rankings, which solves these issues.

  The \emph{agreement} at depth $n$ quantifies
  how many documents appear in the top-$n$ of both measures
  when ranking the documents $\vX$.
  Given two measures of similarity $k_a$ and $k_b$, the symmetric agreement $A_n(k_a, k_b)$ is
  defined as
  
  \begin{align}\label{eq-agreement}
    A_{j,n}(k_a, k_b) =
    \frac{
      \left|
        \{i:r_{j,i}(k_a) \leq n\}
        \cap
        \{i:r_{j,i}(k_b) \leq n\}
      \right|
    }{
      n
    }.
  \end{align}
  The agreement may also be computed at \emph{asymmetric} depths, i.e.,
  \begin{align}\label{eq-asymm-agreement}
    A_{j,(m,M)}(k_a, k_b) =
    \frac{
      \left|
        \{i:r_{j,i}(k_a) \leq m\}
        \cap
        \{i:r_{j,i}(k_b) \leq M\}
      \right|
    }{
      m
    },
  \end{align}
  where $m < M$, counts whether the top-$m$ according to $k_a$
  are contained in the top-$M$ according to $k_b$.
  This can be used to identify cases where, for example,
  the top-$m$ according to one method may be contained in the top-$M$ of the other, but not
  vice-versa.

  Rank-biased overlap~\citep{webber2010similarity} can be used to obtain a more refined similarity
  between rankings by weighting the agreement values up to some depth $d$, with
  \begin{align}
    \text{RBO}_j(k_a, k_b, p) =
      \frac{1-p}{p} \left(
        \sum_{m=1}^{N} A_{j,m}(k_a, k_b) \cdot p^d
        + \frac{A_{j,N}(k_a, k_b) \cdot p^N}{1-p}
      \right),\label{eq:rbo}
  \end{align}
  where $p\in (0, 1]$ is the \emph{persistence}, a hyper-parameter to indicate the top-ranks that
  should be considered.
  The persistence $p = 1 - \frac{1}{d}$ indicates up to which depth $d$
  we care about the ranking, for example, looking at the top $d=100$ results in a
  persistence of $0.99$.

\FloatBarrier
\section{Experimental Details}\label{app:details}
\FloatBarrier

We used the following models with checkpoints as provided by their respective maintainers on Huggingface:
\vspace{-6pt}
\begin{itemize}
  \setlength{\itemsep}{-0.4pt}
  \item \texttt{allenai/OLMo-2-0425-1B-Instruct}~\citep{walsh2025olmo}
  \item \texttt{google/gemma-3-1b-it}~\citep{gemma2025gemma}
  \item \texttt{meta-llama/Llama-3.2-1B}~\citep{grattafiori2024llama}
  \item \texttt{Qwen/Qwen3-1.7B}~\citep{yang2025qwen}
  \item \texttt{team-approx-bayes/gpt2-medium}~\citep{shen2024variational}
  \item \texttt{team-approx-bayes/gpt2-small}~\citep{shen2024variational}
\end{itemize}
\vspace{-6pt}
We use the Kronfluence Python library~\citep{grosse2023studying} to compute EK-FAC, Diagonal
(EK-FAC), and Identity of the complete query (prompt and response).

We used a random subset 100,000 examples of DCLM~\citep{li2024datacomplm},
where we truncate each document to 300 tokens. The resulting actual text length depends on the
tokenizer of the respective models.
The responses of each LLM were limited to generate a maximum of 300 tokens.
We completed the responses without using a chat template.

We used the NLTK~\citep{bird2004nltk} \texttt{word\_tokenize} function to tokenize all queries and
training documents for BM25. We tried different tokenizers using normalization and word-boundary
splitting, as well as using respective LLM tokenizers, for which we obtained similar results.

We used NVIDIA V100 GPUs with 32GB memory to compute the influence,
which required about 16 hours to compute the token-wise influence scores on all 100,000 training
documents on OLMo-2-1B per query running on a single GPU.

\FloatBarrier
\section{List of Used Prompts}
\FloatBarrier

For all experiments, we generated 100 prompts to generate responses of the LLMs.
The exact prompts are shown in \cref{tbl-prompts} along with their indices.
These prompts were generated using Gemini.
We manually verified their diversity of topics and ambiguity.
For instance, the prompt shown in~\Cref{fig:fig2} has index 14.

\begin{table}[h]
\caption{List of the prompts completed by all LLMs with their respective indices.}
\label{queries}
\label{tbl-prompts}
\begin{center}
  \footnotesize
  \setlength{\arraycolsep}{2pt}
  \setlength{\tabcolsep}{4pt}
  \begin{tabular}{|r|l|r|l|}
    \specialrule{\heavyrulewidth}{0pt}{0pt}
    \rowcolor{gray!20}
    \textbf{Index} & \textbf{Prompt} & \textbf{Index} & \textbf{Prompt}\\
    \specialrule{\lightrulewidth}{0pt}{0pt}
    0 & The color of the sky today is                 & 50 & The ancient castle stood tall, overlooking\\
    1 & A cat's favorite activity is usually          & 51 & She felt a surge of excitement when\\
    2 & Water boils at a temperature of               & 52 & The forest was dark and silent, except for\\
    3 & The capital city of France is                 & 53 & He knew, deep down, that the secret was\\
    4 & In the morning, I like to drink               & 54 & It was a cold, crisp morning as the sun\\
    5 & Seven plus five equals                        & 55 & Laughter is often considered the best\\
    6 & An apple is a type of                         & 56 & The smell of freshly baked bread reminded him of\\
    7 & The largest ocean on Earth is                 & 57 & Regret washed over her when she realized\\
    8 & My favorite kind of music is                  & 58 & Hope flickered in the darkness like\\
    9 & A bicycle has two                             & 59 & The antagonist's motivation was rooted in\\
    10 & Before going to bed, one should always        & 60 & Furthermore, it must be acknowledged that\\
    11 & When the phone rang, she quickly              & 61 & Consequently, the team decided to\\
    12 & He decided to walk to the store because       & 62 & Despite the inherent risks, the explorer\\
    13 & If you mix blue and yellow, you get           & 63 & Subsequently, the findings were published in\\
    14 & To open the locked door, you'll need          & 64 & Conversely, the smaller sample size indicated\\
    15 & The train was delayed due to                  & 65 & In essence, the philosophical debate revolves around\\
    16 & We celebrated the victory by                  & 66 & To illustrate this phenomenon, consider the case of\\
    17 & Suddenly, the lights went out, causing        & 67 & Historically, the region was known for its\\
    18 & She learned to play the guitar so she could   & 68 & On the one hand, the investment seems sound; on the other\\
    19 & After the long hike, they felt                & 69 & Ultimately, the decision will depend on\\
    20 & If I had a superpower, it would be            & 70 & The key to successful gardening is\\
    21 & Even though it was raining, we still          & 71 & A famous quote by Shakespeare is\\
    22 & Whenever he sees a dog, he always             & 72 & My greatest fear has always been\\
    23 & Unless you bring a jacket, you might          & 73 & The ingredients for making a pizza are\\
    24 & Since the cookies were burned, we had to      & 74 & Learning a new language can be difficult but\\
    25 & Provided the weather is nice, the event will  & 75 & The invention of the printing press revolutionized\\
    26 & No matter how hard she tried, she couldn't    & 76 & You should never underestimate the power of\\
    27 & Only if you finish your homework can you      & 77 & The library smelled like old paper and\\
    28 & As soon as the bell rings, the students       & 78 & An important rule in programming is to\\
    29 & Although the book was long, it was still      & 79 & If you look closely, you will notice\\
    30 & Could you please hand me                      & 80 & The mysterious object floated silently in\\
    31 & I wonder why the bird is                      & 81 & Without warning, a portal opened, revealing\\
    32 & Tell me the story of                          & 82 & The hero's journey began with a call to\\
    33 & What is the best way to                       & 83 & Her eyes were the color of\\
    34 & How much time will it take to                 & 84 & The hidden message was encrypted using\\
    35 & Explain the concept of                        & 85 & A talking parrot sat on the shoulder of\\
    36 & Who was the first person to                   & 86 & Legend has it that the lost city of\\
    37 & Do you think it's possible to                 & 87 & The wizard carefully mixed the potions, hoping to\\
    38 & Remember to bring along                       & 88 & The futuristic city was powered entirely by\\
    39 & Please describe the taste of                  & 89 & He carried a leather-bound journal filled with\\
    40 & An algorithm is essentially a set of          & 90 & From a purely economic standpoint, the project is\\
    41 & Data structures are crucial for               & 91 & The challenge lies in accurately measuring\\
    42 & The process of photosynthesis involves        & 92 & A sudden flash of inspiration led to\\
    43 & Artificial intelligence aims to mimic         & 93 & When a person is nervous, they might\\
    44 & In geometry, a circle is defined as           & 94 & It is generally agreed that exercise is beneficial for\\
    45 & Gravity is the force that                     & 95 & The most crucial part of the machine is the\\
    46 & The smallest unit of matter is the            & 96 & This phenomenon occurs chiefly when\\
    47 & Renewable energy sources include              & 97 & The little engine that could famously said\\
    48 & Quantum mechanics deals with                  & 98 & Every journey begins with a single\\
    49 & The purpose of a compiler is to               & 99 & Let's start the meeting by discussing\\
    \specialrule{\heavyrulewidth}{0pt}{0pt}
  \end{tabular}
\end{center}
\end{table}

\FloatBarrier
\section{Complete Results}
\FloatBarrier

\begin{figure}[h]
      \centerline{
        \includegraphics{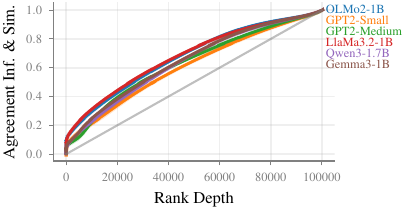}
      }
      \caption{
        Agreement between BM25 and KFAC for various LLMs.%
        The steep ascend of the curves close to rank 1
        indicates some agreement in the top ranks.
        The variance is shown over 100 queries.
          The bottom gray line is the expected agreement for unrelated rankings.
          The plot in the main manuscript including only 20,000 ranks can be found in~\cref{fig-symm}.
      }
      \label{fig-symm-full}
  \end{figure}

\begin{figure*}[t]
    \begin{center}
      \centerline{
        \includegraphics{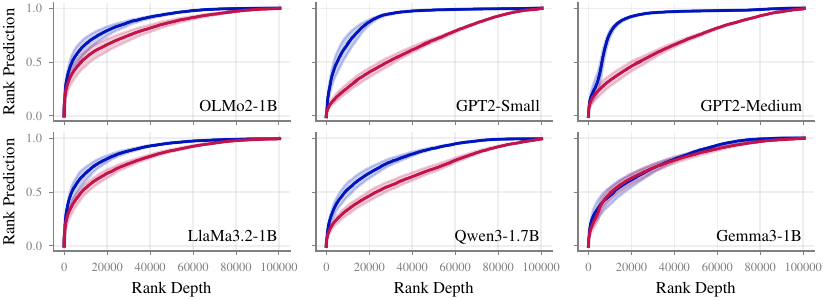}
        \llap{
          \includegraphics{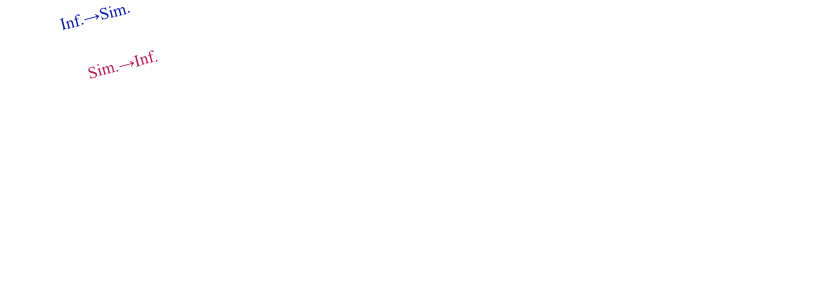}
        }
      }
      \label{fig-asymm-full}
      \caption{
        Results suggesting that Inf is better at predicting Sim than vice versa.
        Rank prediction of the top-100 of Sim (BM25) using Inf (KFAC),
        and vice-versa for six LLMs.
        For the limited 20,000 ranks on only four LLMs, see \cref{fig-asymm}.
      }
    \end{center}
  \end{figure*}

We show the complete results for the experiments conducted in the main manuscript.
Specifically, the full versions of \cref{fig-symm,fig-asymm} are
\cref{fig-symm-full,fig-asymm-full}.

The p-values for the statistical significance of Inf$\rightarrow$Sim predicting more top-ranks
than Sim$\rightarrow$Inf for all rank depths are shown in \cref{fig-p-value} for OLMo2-1B.
Additionally, the rank-biased overlap for all 100 queries on OLMO-2-1B with a persistence of 0.99 (in
other words, for the top-100) are shown in \cref{fig-query-rbo}.
The same figures for all the other models are found in the following figures:
\vspace{-6pt}
\begin{itemize}
  \setlength{\itemsep}{-0.4pt}
  \item \emph{GPT2-Small} in~\cref{fig-addmodel-gpt2-124m}
  \item \emph{GPT2-Medium} in~\cref{fig-addmodel-gpt2-355m}
  \item \emph{LlaMa3.2-1B} in~\cref{fig-addmodel-llama3.2-1b}
  \item \emph{Qwen3-1.7B} in~\cref{fig-addmodel-qwen3-1.7b}
  \item \emph{Gemma3-1B} in~\cref{fig-addmodel-gemma3-1b}
\end{itemize}
\vspace{-6pt}
The gap between \infsim~and \siminf~varies between models, where it appears the widest
overall for
\emph{Qwen3-1.7B} in~\cref{fig-addmodel-qwen3-1.7b},
and the narrowest for
\emph{Gemma3-1B} in~\cref{fig-addmodel-gemma3-1b}.
Here, Gemma3-1B shows a bit of a different behavior, where the predictiveness of \siminf~catches up
to \infsim~in intermediate ranks between 5000 and 6000.
However, it still holds in the top-ranks which are of primary interest.

  \begin{figure}[t]
    \begin{subfigure}{0.49\textwidth}
      \begin{center}
        \centerline{
          \includegraphics{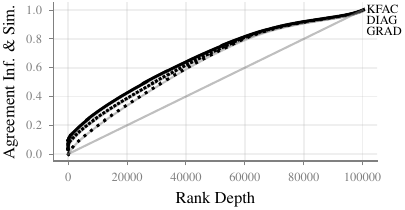}
        }
        \caption{}
        \label{fig-symm-multiple-full}
      \end{center}
    \end{subfigure}
    \hfill
    \begin{subfigure}[b]{0.49\textwidth}
      \begin{center}
        \centerline{
          \includegraphics{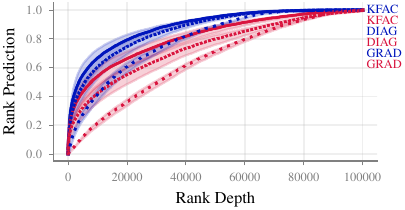}
          \llap{
            \includegraphics{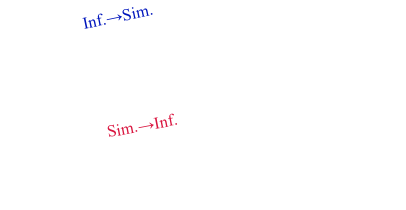}
          }
        }
        \caption{}
        \label{fig-asymm-multiple-full}
      \end{center}
    \end{subfigure}
    \vskip -1em
    \caption{
      Results showing the consistency of our observations across various influence estimators on
      OLMo2-1B.
      (a) Agreement between Sim (BM25) and Inf (various).
      The variance is shown over 100 queries.
        The bottom gray line is the expected agreement for unrelated rankings.
      (b) Rank prediction of the top-100 of Sim (BM25) using Inf (various) and vice-versa.
      The variances are hidden to improve readability.
      For only 20,000 ranks and no variances, see \cref{fig-symm-multiple,fig-asymm-multiple}.
    }
  \end{figure}

    \begin{figure}[h]
      \centering
      \begin{subfigure}[b]{0.30\textwidth}
        \centerline{
          \includegraphics{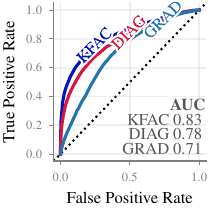}
        }
        \caption{}
      \end{subfigure}
      \begin{subfigure}[b]{0.30\textwidth}
        \centerline{
          \includegraphics{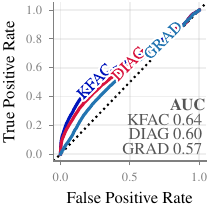}
        }
        \caption{}
      \end{subfigure}
      \caption{
          Pre-filtering using Sim and refining using Inf (a) and vice versa (b), evaluated using the
          receiver-operator characteristic on OLMo2-1B for various influence estimators.
          See experiment step H for details.
        }
      \label{fig-auc-if}
    \end{figure}

  \begin{figure}[h]
    \begin{subfigure}[b]{0.48\textwidth}
      \centerline{
        \includegraphics{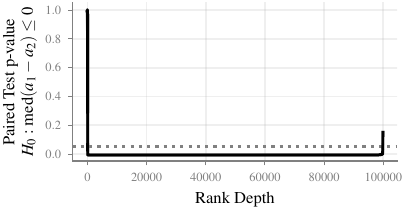}
      }
      \caption{}
      \label{fig-p-value}
    \end{subfigure}
    \hfill
    \begin{subfigure}[b]{0.48\textwidth}
      \begin{center}
        \centerline{
          \includegraphics{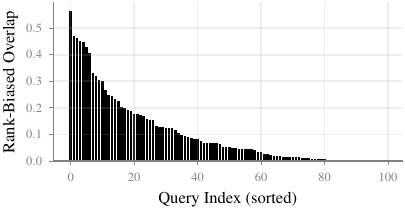}
        }
        \caption{}
        \label{fig-query-rbo}
      \end{center}
    \end{subfigure}
    \caption{
      Paired test (a) and per-query rank-biased overlap for OLMo2-1B.
      (a) The p-value at rank depth $d$ for the null hypothesis that Inf (KFAC) at depth $d$ ranks
      the top 100 according to Sim (BM25)~inferior or equal compared to how BM25 at depth $d$
      ranks the top 100 according to Inf.
      (b) Rank-biased overlap with persistence $0.99$ (i.e., top-100) for each of 100 queries between
      Sim (BM25) and Inf (K-FAC) sorted descending by the overlap.
      A coefficient of $1.0$ indicates a perfect match between the rankings, and a
      coefficient of $0.0$ indicates no match.
    }
  \end{figure}

\newcommand\makemodelfig[2]{%
  \begin{figure}[ht]
    \begin{center}
      \begin{subfigure}[b]{0.48\textwidth}
        \centering
        \includegraphics{{#1--asymm-agreement-wilcoxon-bm25-ekfac}.pdf}
        \caption{}
      \end{subfigure}
      \hfill
      \begin{subfigure}[b]{0.48\textwidth}
        \centering
        \includegraphics{{#1--rbo-queries-bm25-ekfac}.pdf}
        \caption{}
      \end{subfigure}
    \end{center}
      \caption{
        Paired test (a) and per-query rank-biased overlap (b) between Sim (BM25) and Inf
        (KFAC) on \emph{#2}.
      }
    \label{fig-addmodel-#1}
  \end{figure}
}

\makemodelfig{gpt2-124m}{GPT2-Small}
\makemodelfig{gpt2-355m}{GPT2-Medium}
\makemodelfig{llama3.2-1b}{LlaMa3.2-1B}
\makemodelfig{qwen3-1.7b}{Qwen3-1.7B}
\makemodelfig{gemma3-1b}{Gemma3-1B}

\FloatBarrier
\section{Additional Token-wise Visualizations}\label{app-qualitative}
\FloatBarrier

In
\cref{%
fig-sentencetoken-14-bm25,
fig-sentencetoken-14-ekfac,%
fig-sentencetoken-47-bm25,%
fig-sentencetoken-47-ekfac,%
fig-sentencetoken-79-bm25,%
fig-sentencetoken-79-ekfac,%
fig-sentencetoken-0-bm25,%
fig-sentencetoken-0-ekfac,%
fig-sentencetoken-11-bm25,%
fig-sentencetoken-11-ekfac,%
fig-sentencetoken-20-bm25,%
fig-sentencetoken-20-ekfac,%
fig-sentencetoken-24-bm25,%
fig-sentencetoken-24-ekfac,%
fig-sentencetoken-30-bm25,%
fig-sentencetoken-30-ekfac,%
fig-sentencetoken-94-bm25,%
fig-sentencetoken-94-ekfac%
}
we list the full documents for which we conducted the qualitative analysis in \cref{sec-methods}.

\newcommand\maketokenfig[3]{%
  \begin{figure*}[ht]
    \begin{center}
      \centerline{
        \includegraphics{sentencetoken-dclm-bm25-ekfac-#1.pdf}
      }
      \caption{
        Top-3 training documents according to #2 on OLMo2-1B
        for prompt: `\emph{#3...}'.
      }
      \label{fig-sentencetoken-#1}
    \end{center}
  \end{figure*}
}
\newcommand\maketokenfigpairs[2]{%
  \maketokenfig{#1-bm25}{data-similarity (BM25)}{#2}
  \maketokenfig{#1-ekfac}{data-influence (EK-FAC)}{#2}
}

\maketokenfigpairs{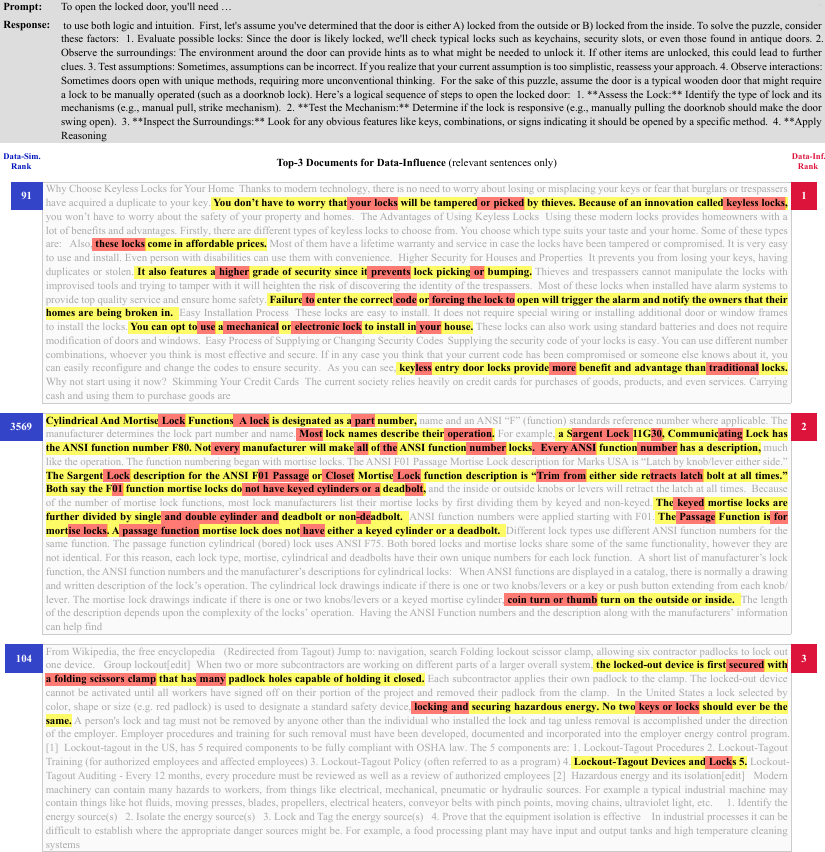}{To open the locked door, you'll need}
\maketokenfigpairs{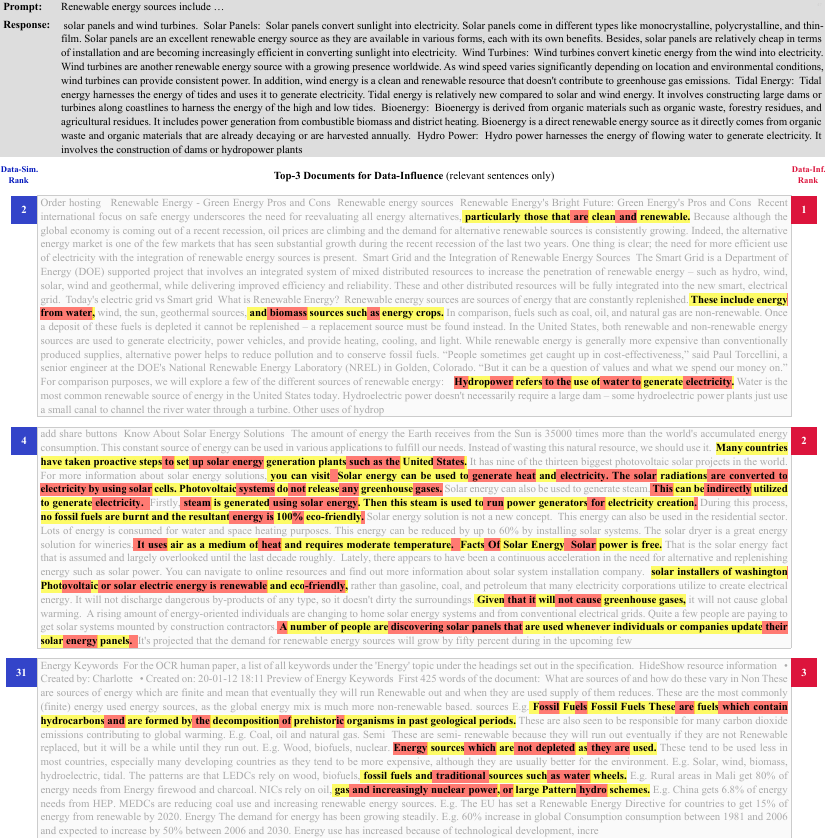}{Renewable energy sources include}
\maketokenfigpairs{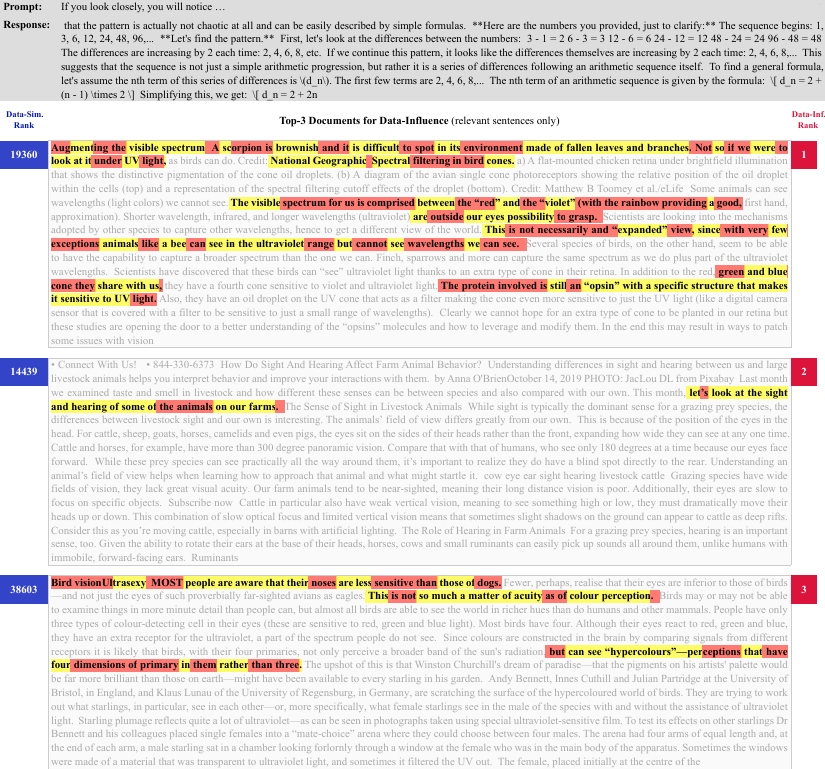}{If you look closely, you will notice}

\maketokenfigpairs{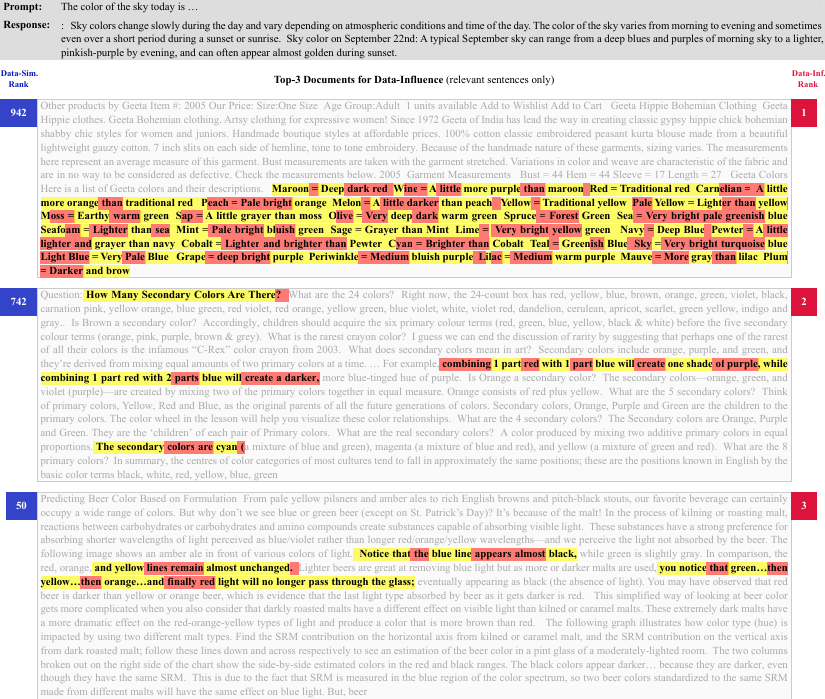}{The color of the sky today is}
\maketokenfigpairs{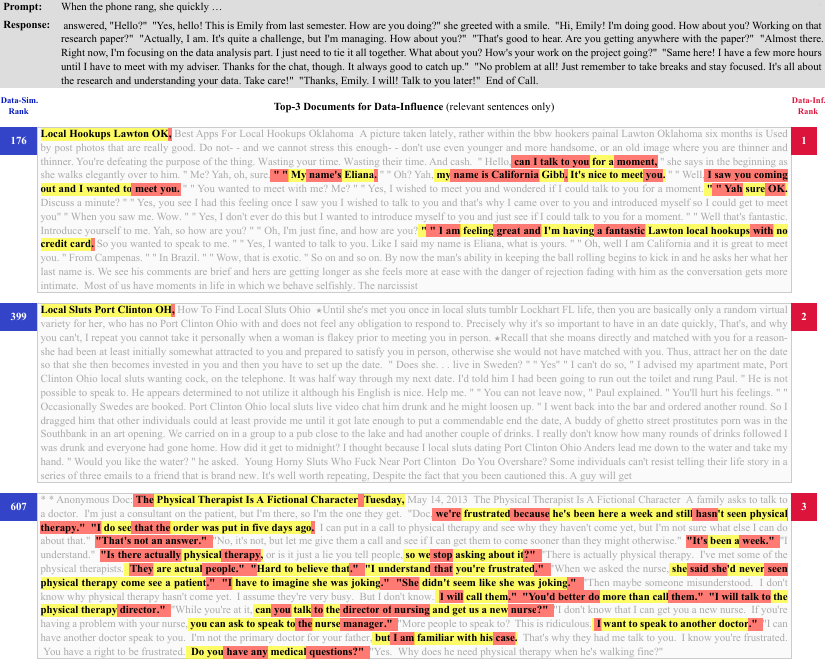}{When the phone rang, she quickly}
\maketokenfigpairs{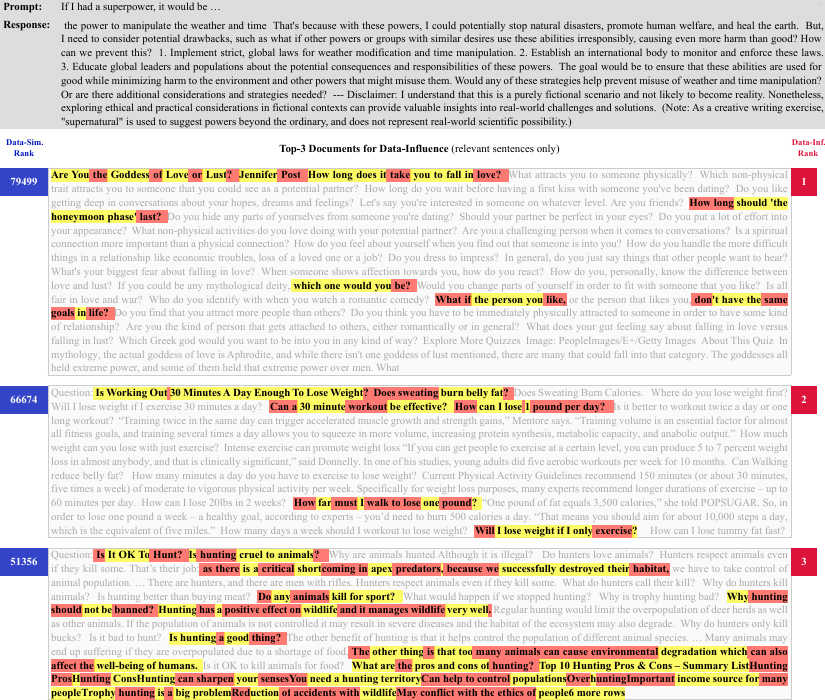}{If I had a superpower, it would be}
\maketokenfigpairs{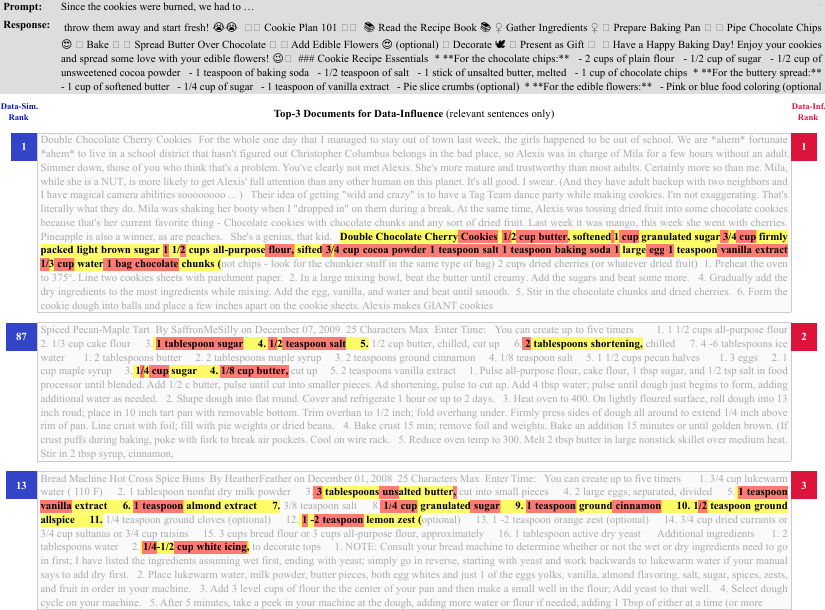}{Since the cookies were burned, we had to}
\maketokenfigpairs{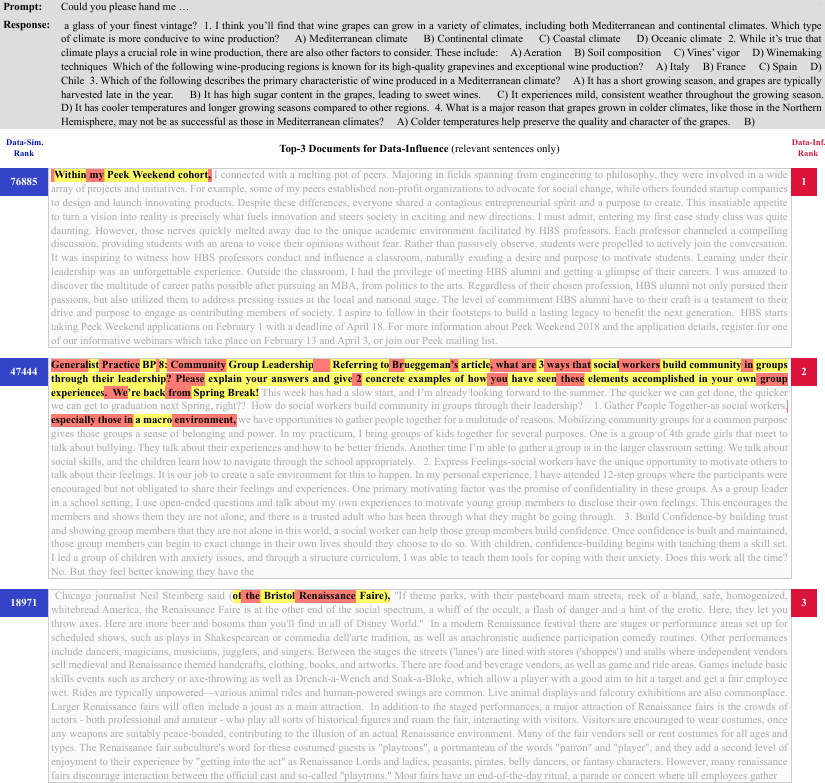}{Could you please hand me}
\maketokenfigpairs{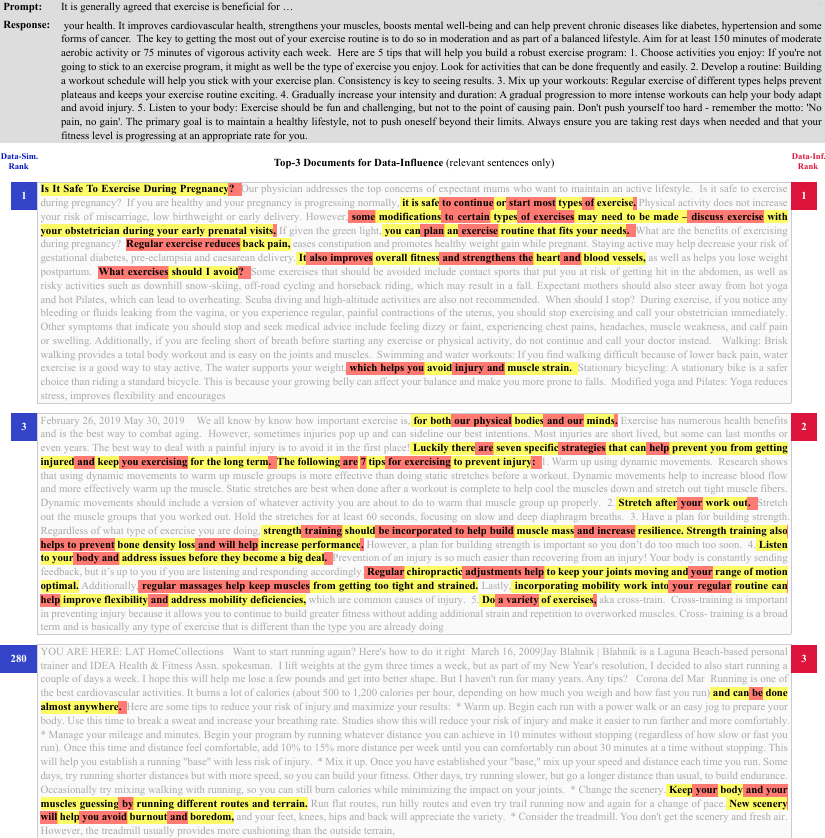}{It is generally agreed that exercise is beneficial for}

\FloatBarrier

\end{document}

%% file: defns.tex
\newcommand{\minus}{-}

\newcommand{\influence}{\mathcal{I}}

\newcommand{\vparam}{\boldsymbol{\theta}}

\newcommand\cut[1]{}

\newcommand{\squishlist}{
   \begin{list}{$\bullet$}
    { \setlength{\itemsep}{0pt}      \setlength{\parsep}{3pt}
      \setlength{\topsep}{3pt}       \setlength{\partopsep}{0pt}
      \setlength{\leftmargin}{1.5em} \setlength{\labelwidth}{1em}
      \setlength{\labelsep}{0.5em} } }

\newcommand{\squishlisttwo}{
   \begin{list}{$\bullet$}
    { \setlength{\itemsep}{0pt}    \setlength{\parsep}{0pt}
      \setlength{\topsep}{0pt}     \setlength{\partopsep}{0pt}
      \setlength{\leftmargin}{2em} \setlength{\labelwidth}{1.5em}
      \setlength{\labelsep}{0.5em} } }

\newcommand{\squishend}{
    \end{list}  }

{}
{}
{}
{}
\newcommand{\real}{\mbox{$\mathbb{R}$}}

\newcommand{\myvec}[1]{\mbox{$\mathbf{#1}$}}
\newcommand{\myvecsym}[1]{\mbox{$\boldsymbol{#1}$}}

\newcommand{\vphi}{\mbox{$\myvecsym{\phi}$}}

\newcommand{\vd}{\mbox{$\myvec{d}$}}

\newcommand{\vf}{\mbox{$\myvec{f}$}}
\newcommand{\vg}{\mbox{$\myvec{g}$}}

\newcommand{\vq}{\mbox{$\myvec{q}$}}

\newcommand{\vw}{\mbox{$\myvec{w}$}}

\newcommand{\vx}{\mbox{$\myvec{x}$}}

\newcommand{\vH}{\mbox{$\myvec{H}$}}

\newcommand{\vX}{\mbox{$\myvec{X}$}}

\newcommand{\calA}{\mbox{${\cal A}$}}

\newcommand{\calD}{\mbox{${\cal D}$}}

\newcommand{\be}{\begin{equation}}
\newcommand{\ee}{\end{equation}}
\newcommand{\bea}{\begin{eqnarray}}
\newcommand{\beaa}{\begin{eqnarray*}}
\newcommand{\eeaa}{\end{eqnarray*}}
